# Genetic Algorithm for the 0/1 Multidimensional Knapsack Problem

Shalin Shah
sshah100@jhu.edu
shah.shalin@gmail.com
Johns Hopkins University

The 0/1 multidimensional knapsack problem is the 0/1 knapsack problem with m constraints which makes it difficult to solve using traditional methods like dynamic programming or branch and bound algorithms. We present a genetic algorithm for the multidimensional knapsack problem with Java and C++ code that is able to solve publicly available instances in a very short computational duration. Our algorithm uses iteratively computed Lagrangian multipliers as constraint weights to augment the greedy algorithm for the multidimensional knapsack problem and uses that information in a greedy crossover in a genetic algorithm. The algorithm uses several other hyperparameters which can be set in the code to control convergence. Our algorithm improves upon the algorithm by Chu and Beasley in that it converges to optimum or near optimum solutions much faster.

*1. Introduction*

Solving the multidimensional knapsack problem using branch and bound or dynamic programming is difficult. Because of the multiple constraints, it is also difficult to obtain a good approximation to the solution such as a greedy algorithm. However, it is possible to use the greedy algorithm as part of a genetic algorithm, and our results show that it works really well. Not only is our algorithm able to exceed the greedy estimate, but for most problem instances, it is able to find the optimum solution. Our algorithm is similar to [2] which uses greedy crossover for the 0/1 knapsack problem. Since the multidimensional knapsack problem has multiple constraints, we assign a weight to each constraint using iteratively computed Lagrangian multipliers. This is similar to the approach in [1] which uses surrogate multipliers. The difference is that we use the multipliers in a greedy crossover which is highly constructive and can find optimum solutions much quicker.

*2. Problems and Background*

The 0/1 multidimensional knapsack problem is the 0/1 knapsack problem with m constraints which makes it difficult to solve using traditional methods. The 0/1 multidimensional knapsack problem can be stated as: Given n objects each with a value $v_i$ and m constraints (or knapsacks) each with a capacity constraint $c_j$, maximize the value such that each of the m constraints are satisfied. Each of the m constraints have *i* weights associated with it. This makes it a general 0/1 integer programming problem.

Maximize: $\sum_{i=1}^{n} x_i v_i$

Such that: $\sum_{i=1}^{n} x_i w_{i,j} \leq c_j$ , j: 1...m

$x_i \in \{0,1\}$

Our algorithm generates the initial population with the probability of choosing an object 0.5. In a problem with n objects, $n/2$ are chosen on an average. The higher this probability, the faster the algorithm converges; however, the higher this probability, the more are the chances that the algorithm will converge around the greedy estimate. This way of generating the initial population introduces a lot of invalid solutions (noise) into the population. (Strategies for initial population generation are discussed in [5]). To compensate for invalid solutions, we investigated the use of a highly constructive greedy crossover. The greedy crossover takes the objects with the best utility ratio from parents and constructs one offspring such that it is always a valid solution.

We use Lagrangian multipliers to augment the utility ratio for the multidimensional knapsack problem according to the following steps:

1. For each object and for each constraint (for that object) the weight (constraint) value is multiplied with the corresponding Lagrangian multiplier $l_j$ and the sum of these values is obtained.
2. The value obtained in step 1 is then divided by the number of constraints (optional step).
3. Then, the ratio of the value (profit) and the value obtained in step 2 is obtained which is the profit-weight ratio for that object

$$ratio_i = v_i / ((\sum_{j=1}^{m} l_j * w_{i,j}) / m)$$

Where $l_j$ is the j[th] Lagrangian multiplier and *m* is the number of constraints. The greedy crossover simply takes objects from the two parents in non-increasing order of the ratio and constructs one offspring such that it satisfies all constraints.

*3. Software Framework*

Our code is written in Java and C++. Any JDK compiler and gcc should work. Benchmark instances that we use in this paper are available at [3]. Our code is available at [4]. Also see [8]. The code requires a data.DAT file in the directory in which the executable resides. Change Constants.java to increase or decrease the number of generations (Constants.GENERATIONS). Change the data file name in Constants.java to use some other Weing/Weish/Sento file. Change the DataProcessor in GeneticAlgorithm.java to ORLIB if required.

*4. Implementation and Empirical Results*

We ran our algorithm in publicly available instances. Some results are shown below. More results are available in our git repository [4]. Our algorithm is able to solve most instances completely, reaching the global optimum.

Table-1: Our algorithm applied to some benchmark instances [3].

| Instance | m | n | Solves Completely | Time (mean) | % gap |
|---|---|---|---|---|---|
| Sento1 | 30 | 60 | 13\20 | 4.8 seconds | 0 |
| Sento2 | 30 | 60 | 20\20 | 0.2 seconds | 0 |
| Weing1 | 2 | 28 | 11\20 | 3.5 seconds | 0 |
| Weing5 | 2 | 28 | 20\20 | 0.6 seconds | 0 |
| Weing7 | 2 | 105 | 1\20 | 27.4 seconds | 0 |
| Weing8 | 2 | 105 | 11\20 | 2 seconds | 0 |
| Weish05 | 5 | 30 | 20\20 | 0.02 seconds | 0 |
| Weish10 | 5 | 50 | 16\20 | 0.1 seconds | 0 |
| Weish15 | 5 | 60 | 16\20 | 1.2 seconds | 0 |
| Weish20 | 5 | 70 | 10\20 | 10 seconds | 0 |
| Weish25 | 5 | 80 | 10\20 | 1.9 seconds | 0 |
| Weish30 | 5 | 90 | 13\20 | 3.6 seconds | 0 |

$m$ is the number of constraints and $n$ is the number of objects.

## 5. Illustrative Examples

Our algorithm can be applied to any 0/1 integer programming problem and the utility ratio is general enough for most types of inequality constraints. We haven't tried running our algorithm in equality constraints though it should be trivial. Our algorithm is fast and can find optimum solutions quite fast. Instances with larger number of objects are shown in our git repository [4][8].

## 6. Conclusions

Traditional evolutionary algorithms are more suitable for problems in which domain specific knowledge is not available. For problems with partial knowledge of the domain, a genetic algorithm, which uses this domain knowledge, is more likely to succeed, as the results clearly indicate. A good search algorithm should be global in nature with a heuristic introduced to give constructive direction to the algorithm. We introduced a new technique of greedy crossover; it forms the core of our genetic algorithm. As table-1 shows, our algorithm is able to solve to optimality, all of the instances in a short amount of time. Some problems like Weing7 are harder. Future work could be to run the algorithm on larger instances for which optimum solutions are available. Our algorithm is trivially parallelizable and future work could be to implement the algorithm on Apache Spark or Map-Reduce.

**Table 2 – Code metadata**

| Nr | Code metadata description | https://github.com/shah314/gamultiknapsack |
|---|---|---|
| C1 | Current Code version | V1.7 |
| C2 | Permanent link to code / repository used of this code version | https://github.com/shah314/gamultiknapsack |
| C3 | Legal Code License | MIT License |
| C4 | Code Versioning system used | Git |
| C5 | Software Code Language used | Java and C++ |
| C6 | Compilation requirements, Operating environments & dependencies | **Java implementation**<br>I tested the code using JDK 1.8, but any JDK should work fine. If the code does not compile, please open an issue. Compile the Java code and then run GeneticAlgorithm.<br><br>javac *.java<br>java GeneticAlgorithm filename format<br><br>(The file name contains the instance in weing or orlib format)<br>(The format is either weing or orlib)<br><br>Example:<br>java GeneticAlgorithm data.DAT weing<br><br>**C++ implementation**<br>The code was tested on a Mac with gcc version 8, downloaded using homebrew. If the code does not compile, please open an issue. Compile the C++ code and then run the executable.<br><br>g++ cmultiknapsack.cpp<br>./a.out filename format<br><br>(The file name contains the instance in weing or orlib format)<br>(The format is either weing or orlib)<br><br>Example:<br>./a.out data.DAT weing<br><br>(Please remove all comments and other extraneous text from data.DAT) |
| C7 | If available Link to developer documentation / manual | https://github.com/shah314/gamultiknapsack<br>See [8] |
| C8 | Support email for questions | *shah.shalin@gmail.com* |